\documentclass[10pt,twocolumn,letterpaper]{article}

\usepackage{wacv}
\usepackage{times}
\usepackage{epsfig}
\usepackage{graphicx}
\usepackage{amsmath}
\usepackage{amssymb}
\usepackage{booktabs}
\usepackage{multirow}
\usepackage{caption}
\usepackage{setspace}
\usepackage{mathtools}
\usepackage{textcomp}
\usepackage[accsupp]{axessibility}  % Improves PDF readability for those with disabilities.
\usepackage{balance}

\def\wacvPaperID{588} % Enter the WACV Paper ID here

\wacvfinalcopy % *** Uncomment this line for the final submission

\ifwacvfinal
\usepackage[breaklinks=true,bookmarks=false]{hyperref}
\else
\usepackage[pagebackref=true,breaklinks=true,colorlinks,bookmarks=false]{hyperref}
\fi

\usepackage{fancyhdr}

\begin{document}

\title{Creating and Reenacting Controllable 3D Humans with Differentiable Rendering\vspace*{-0.1cm}}

\author{Thiago L. Gomes$^1$ \hspace{2cm} Thiago M. Coutinho$^1$ \hspace{2cm}  Rafael Azevedo$^1$ \hspace{2cm}\\ 
	Renato Martins$^{2}$ \hspace{2cm} Erickson R. Nascimento$^1$ \\
	\hspace{-0.5cm} $^1$Universidade Federal de Minas Gerais (UFMG), Brazil \hspace{0.3cm} $^2$ Université Bourgogne Franche-Comté, France\\
	{\tt\small \{thiagoluange,thiagocoutinho,rafaelvieira\}@dcc.ufmg.br, \vspace*{-0.1cm}} \\ 
	{\tt\small renato.martins@u-bourgogne.fr, erickson@dcc.ufmg.br}\vspace*{-0.2cm}
}

\maketitle
%\ifwacvfinal\thispagestyle{empty}\fi
\thispagestyle{fancy}
\fancyhf{}
\chead{{To appear in Proceedings of the IEEE Winter Conference on Applications of Computer Vision (WACV) 2022 \\ The final publication will be available soon.}}
\cfoot{1}

%%%%%%%%% ABSTRACT
\begin{abstract}
   This paper proposes a new end-to-end neural rendering architecture to transfer appearance and reenact human actors. Our method leverages a carefully designed graph convolutional network (GCN) to model the human body manifold structure, jointly with differentiable rendering, to synthesize new videos of people in different contexts from where they were initially recorded. Unlike recent appearance transferring methods, our approach can reconstruct a fully controllable 3D texture-mapped model of a person, while taking into account the manifold structure from body shape and texture appearance in the view synthesis. Specifically, our approach models mesh deformations with a three-stage GCN trained in a self-supervised manner on rendered silhouettes of the human body. It also infers texture appearance with a convolutional network in the texture domain, which is trained in an adversarial regime to reconstruct human texture from rendered images of actors in different poses. Experiments on different videos show that our method successfully infers specific body deformations and avoid creating texture artifacts while achieving the best values for appearance in terms of Structural Similarity (SSIM), Learned Perceptual Image Patch Similarity (LPIPS), Mean Squared Error (MSE), and Fréchet Video Distance (FVD). By taking advantages of both differentiable rendering and the 3D parametric model, our method is fully controllable, which allows controlling the human synthesis from both pose and rendering parameters. The source code is available at {\color{magenta}\url{https://www.verlab.dcc.ufmg.br/retargeting-motion/wacv2022}}.\vspace*{-0.4cm}
 \end{abstract}

\section{Introduction}

\begin{figure}[t!]
	\centering
	\includegraphics[width=0.95\linewidth]{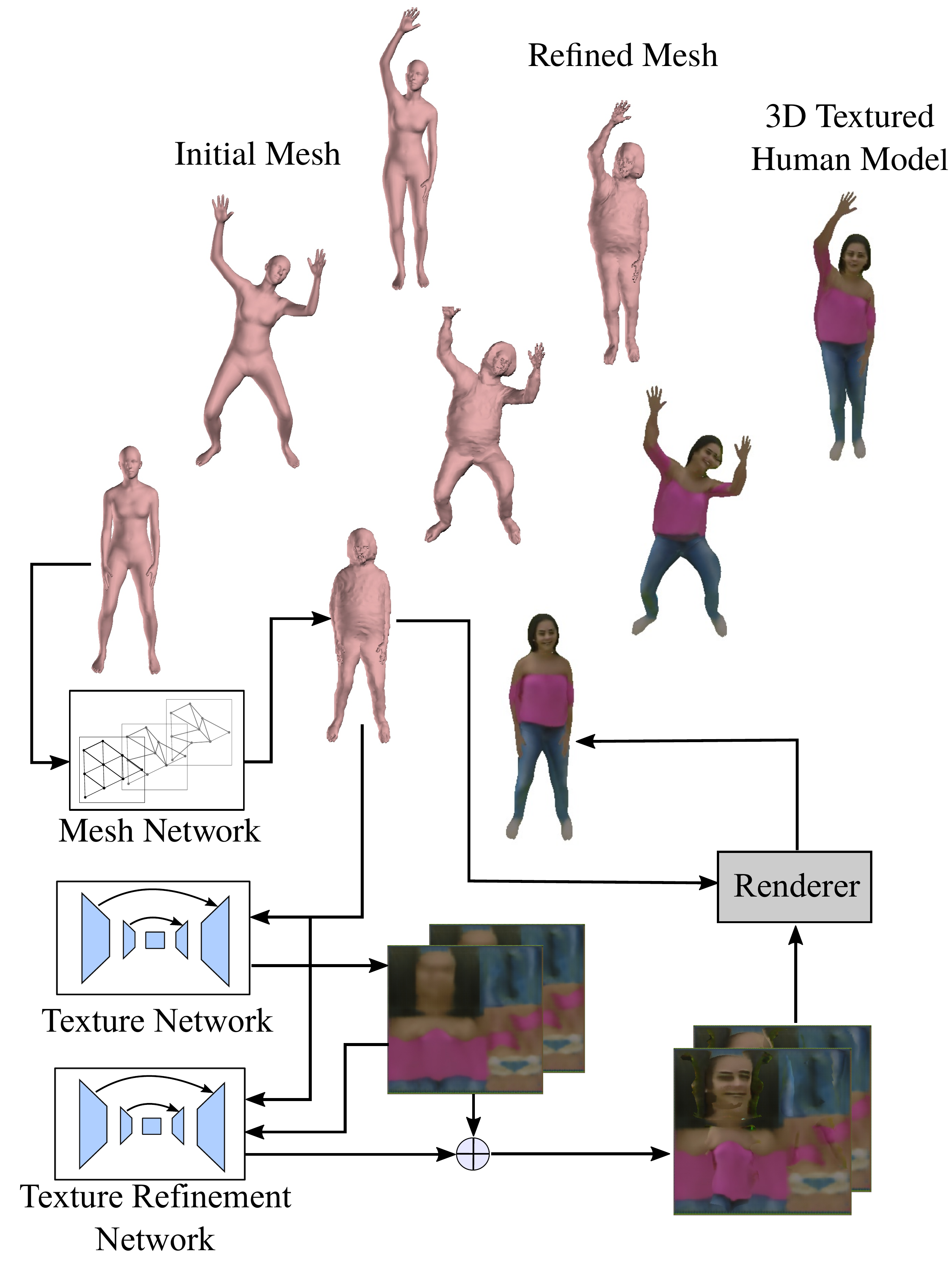}\vspace*{-0.4cm}
	\caption{{\bf 3D texture-mapped human synthesis.} Our method receives a frame of a person, extracts her/his mesh (left side) and outputs a refined 3D shape and appearance representations of people (right side).}
	\label{fig_method2}
	\vspace{-0.3cm}
\end{figure}

%%% Establish the territory %%%
% 1) Deep learning + Rendering = better results

The research in computer vision and graphics communities has made great strides forward in the last two decades, with milestones being passed from semantic segmentation~\cite{minaee2020image}, tridimensional reconstruction~\cite{natsume2019siclope,saito2019pifu,saito2020pifuhd} to synthesis of human motion~\cite{ferreira2020cag,NEURIPS2019_dancing2music}, realistic images~\cite{CycleGAN2017} and videos~\cite{wang2018vid2vid}. As more vision and graphics methods are integrated into new approaches such as differentiable rendering, more systems will be able to achieve high accuracy and quality in different tasks, in particular, generative methods for synthesizing videos with plausible human reenactment.

%%% Establish the niche - what is missing in the field? %%%
% 1) GAN doesn't create 3D and image-based methods demand painstaking designing
% 2) learning based can be improved with more data!

Over the past few years, a remarkable performance of the Generative Adversarial Networks (GANs)~\cite{goodfellow2014generative} has shed a new light for the problem of synthesizing faithful real-world images of humans. %With the success of the GANs, we have witnessed the rise of new applications such as synthesis of human faces~\cite{karras2018progressive}, image-to-image and video-video translation~\cite{pix2pix2017,CycleGAN2017,wang2018vid2vid,Esser_2018_CVPR}, motion synthesis~\cite{ferreira2020cag,NEURIPS2019_dancing2music}, and human retargeting and reenacting~\cite{chan2018dance,wayne2018reenactgan,liu2019neural}, to name a few.
Although GAN-based approaches have been achieving high quality results in generating videos~\cite{chan2018dance,wayne2018reenactgan,liu2019neural} and images~\cite{karras2018progressive,pix2pix2017,CycleGAN2017,wang2018vid2vid,Esser_2018_CVPR} of people, in general, they suffer with the high variability in the poses,  unseen images from viewpoints not present in the training data, and are often limited to reason directly on the projected 2D image domain. Image-based rendering techniques~\cite{Gomes_2020_WACV,gomes2020arxiv}, for their turn, are effective solutions to create 3D texture-mapped models of people, being capable to synthesize images from any arbitrary viewpoint without using large number of images. On the other hand, image-based rendering methods require a painstaking design of the algorithm and are not capable of improving the visual quality of the synthesized images by using more data when available.  

%%% Occupy the niche (i.e., purpose statement) %%%
In this paper, we take a step towards combining learning and image-based rendering approaches in a new end-to-end architecture that synthesizes human avatars capturing both body geometry and texture details. The proposed architecture comprises a graph convolution network (GCN) operating over the mesh with differentiable rendering to achieve high-quality results in image generation of humans. Our approach receives as input a frame of a person, estimates a generic mesh in the desired pose and outputs a representation of their shape and appearance, as shown in Figure~\ref{fig_method2}. The method provides a final 3D representation that is compatible with traditional graphic pipelines. Specifically, our architecture estimates a refined mesh and a detailed texture map to properly represent the person's shape and appearance for a given input frame.

%%% Show to the reader the importance of the niche %%%

While 2D image-to-image translation methods are capable of generating plausible images, many applications such as Virtual Reality (VR) and Augmented Reality (AR)~\cite{Gallala_2019,Chen,minaee2020image} require a fully 3D representation of the person. Additionally, the view-dependence hinders the creation of new configurations of scenes where the avatar can be included. Although view interpolation can be applied to estimate a transition between a pair of camera poses, it may create artifacts and unrealistic results when adding the virtual avatar into the new scene using unseen camera poses. Video-based rendering systems are greatly benefited through the use of realistic 3D texture-mapped models, which make possible the inclusion of virtual avatars using unrestricted camera poses and the automatic modification and re-arrangement of video footage. In addition to being able to render human avatars from different viewpoints, the 3D shape also allows synthesizing new images under different illumination conditions. We experimentally showed the capabilities of our new architecture to estimate realistic 3D texture-mapped models of humans. Experiments on a variety of videos show that our method excels several state-of-the-art methods achieving the best values for appearance in terms of Structural Similarity (SSIM), Learned Perceptual Image Patch Similarity (LPIPS), Mean Squared Error (MSE), and Fréchet Video Distance (FVD). 
\vspace*{-0.35cm}
\paragraph{Contributions.} The main contributions of this paper are threefold: i) a novel formulation for transferring appearance and reenacting human actors that produces a fully 3D representation of the person; ii) a graph convolutional architecture for mesh generation that leverages the human body structure information and keeps vertex  consistency, which results in a refined human mesh model; iii) a new architecture that takes advantages of both differentiable rendering and the 3D parametric model and provides a fully controllable human model, \ie, the user can control the human pose and rendering parameters.
%\vspace*{-0.35cm}

\section{Related Work}

\paragraph{Human animation by neural rendering.} Recently, we witnessed the explosion of neural rendering approaches to animate and synthesize images of people on unseen poses~\cite{wang2018vid2vid,Esser_2018_CVPR,liu2019neural}. According to Tewari~\etal~\cite{tewari2020cgf}, neural rendering is a deep image or video generation approach that enables explicit or implicit control of scene properties. The main difference among these approaches is how the control signal is provided to the network. Lassner~\etal~\cite{Lassner_GeneratingPeople} proposed a GAN called ClothNet. ClothNet produces random people with similar pose and shape in different clothing styles using as a control signal a  silhouette image. In a similar context, Esser~\etal~\cite{Esser_2018_CVPR} used a conditional U-Net to synthesize new 2D human views based on estimated edges and body joint locations. Chan~\etal~\cite{chan2018dance} applied an adversarial training to map a 2D source pose to the appearance of a target subject. Wang~\etal~\cite{wang2018vid2vid} presented a general video-to-video synthesis framework based on conditional GANs, where a 2D dense UV mapping to body surface (DensePose~\cite{Guler2018DensePose}) is used as a control signal. Similarly, Neverova~\etal~\cite{NeverovaGK18} and Sarkar~\etal~\cite{sarkar2020neural} rely on DensePose as input to synthesize new views.  

These image-to-image translation approaches only allow for implicit control by way of representative samples, \ie, they can copy the scene parameters from a reference image/video, but not manipulate these parameters explicitly. To enable the control of the position, rotation, and body pose of a person in a target image/video,  Liu~\etal~\cite{liu2019neural} proposed to use a medium-quality controllable 3D template model of people. In the same line, Liu~\etal~\cite{lwb2019} proposed a 3D body mesh recovery module based on a parametric statistical human body model SMPL~\cite{Loper_2015}, which disentangles human body into joint rotations and shape. Wu~\etal~\cite{wu2020multi} produced photorealistic free-view-video from multi-view dynamic human captures. Although these methods are capable of generating plausible 2D images, they cannot generate a controllable 3D models of people, unlike our approach, which is a desired feature in many tasks such as in rendering engines and games or virtual and augmented reality contexts~\cite{Gallala_2019,Chen,minaee2020image}. 

\vspace*{-0.4cm}\paragraph{3D human pose and mesh reconstruction.} Substantial advances have been made in recent years for human pose and 3D model estimation from still images. %Sigal~\etal~\cite{Sigal} estimate the human body shape by fitting a generative model, the SCAPE~\cite{Anguelov_2005}, to image silhouettes. 
Bogo~\etal~\cite{Bogo_2016} proposed the SMPLify method, a fully automated approach for estimating 3D body shape and pose from 2D joints in images. SMPLify uses a CNN to estimate 2D joint locations and then optimize the re-projection errors of an SMPL body model~\cite{Loper_2015} to these 2D joints. Similarly, Kanazawa~\etal~\cite{kanazawaHMR18} used unpaired 2D keypoint annotations and 3D scans to train an end-to-end network to regress the 3D mesh parameters and the camera pose. Kolotouros~\etal~\cite{kolotouros2019spin} proposed SPIN, an hybrid approach combining the ideas of optimization-based method from~\cite{Bogo_2016} and regression deep network~\cite{kanazawaHMR18} to design an efficient method less sensitive to the optimization initialization, while keeping accuracy from the optimization-based method. %Lin~\etal~\cite{lin2021end} used a transformer encoder to jointly model vertex-vertex and vertex-joint interactions, and outputs 3D joint coordinates and mesh vertices simultaneously. 

Despite remarkable results in pose estimation, these methods are limited to estimate coarse quality generic meshes and textureless characters. %Following the same trend, human mesh reconstruction methods are also increasingly achieving better results as shown in works such as PiFu~\cite{saito2019pifu,saito2020pifuhd}, ARCH~\cite{huang2020arch} or SiCloPe~\cite{natsume2019siclope}. 
%Despite the impressive results, these methods still obtain static 3D character models, which require additional efforts to create animated virtual characters. 
%However, in addition to the requirement that 3D model to contain a skeleton hierarchy and appropriate skin weights, these methods also demand to fit a garment model into a human model in a variety of poses.
Gomes~\etal~\cite{Gomes_2020_WACV,gomes2020arxiv} proposed an image-based rendering technique to create 3D textured models of people synthesized from arbitrary viewpoints. However, their method is not end-to-end and  it is not capable of improving the visual quality of the synthesized images by using information of all available data in the training step. Lazova~\etal~\cite{lazova2019} automatically predict a full 3D textured avatar, including geometry and 3D segmentation layout for further generation control; however, their method cannot predict fine details and complex texture patterns. Methods like PiFu~\cite{saito2019pifu,saito2020pifuhd}, for their turn, are limited to static 3D models per frame. Their mesh represents a rigid body and the users cannot drive the human body to new poses, \ie, it is not possible to make animations where the model raise his/her hands since the arms and hands are not distinguished from the whole mesh. 

\vspace*{-0.4cm}
\paragraph{Graph networks (GCN) and adversarial learning.} GCNs recently emerged as a powerful representation for learning from data lying on manifolds beyond n-dimensional Euclidean vector spaces. They have been widely adopted to represent 3D point clouds such as PointNet~\cite{qi2017pointnet} or Mesh R-CNN~\cite{Gkioxari_2019_ICCV},
%images on non-flat manifolds such as spherical panoramas~\cite{jiang2019spherical,defferrard2020deepsphere}, 
%video forecasting~\cite{ye2019compositional} 
and notably, to model the human body structure with state-of-the-art results in tasks such as human action recognition~\cite{yan2018spatial}, pose estimation~\cite{zhao2019semantic,wang2020motion} and human motion synthesis~\cite{yan2019convolutional,ferreira2020cag,ren_mm_dance}. Often these GCNs have been combined and trained in adversarial learning schemes, as in human motion~\cite{ferreira2020cag}, and pose estimation~\cite{kanazawaHMR18}. Our work leverages these features from GCNs and adversarial training to estimate 3D texture-mapped human models.

%, generative adversarial networks (GAN) have been successfully applied to a myriad of hard problems, notably for the synthesis of new information, such as of images~\cite{pix2pix2017}, motion~\cite{ferreira2020cag}, and pose estimation~\cite{kanazawaHMR18}, to name a few. Gomes~\etal~\cite{Gomes_2020_WACV} synthesize realistic face texture maps from blurring texture maps, demonstrating that GANs can also be used to restore texture maps.
\vspace*{-0.35cm}
\paragraph{Differentiable rendering.} Differentiable renderers (DR) are operators allowing the gradients of 3D objects to be calculated and propagated through images while training neural networks. As stated in~\cite{kato2020differentiable}, DR connects 2D and 3D processing methods and allows neural networks to optimize 3D entities while operating on 2D projections. Loper and Black~\cite{loper2014opendr} introduced an approximate differentiable render which generates derivatives from projected pixels to the 3D parameters. Kato~\etal~\cite{kato2018renderer} approximated the backward gradient of rasterization with a hand-crafted function. Liu~\etal~\cite{liu2019softras} proposed a formulation of the rendering process as an aggregation function fusing the probabilistic contributions of all mesh triangles with respect to the rendered pixels. Niemeyer~\etal~\cite{Niemeyer_2020_CVPR} represented surfaces as 3D occupancy fields and used a numerical method to find the surface intersection for each ray, then they calculate the gradients using implicit differentiation. Mildenhall~\etal~\cite{mildenhall2020nerf} encoded a 3D point and associated view direction on a ray using periodic activation functions, then they applied classic volume rendering techniques to project the output colors and densities into an image, which is naturally differentiable. More recently, techniques~\cite{neural-actor,pumarola2020d} based on neural radiance field (NeRF) learning~\cite{nerf} are being proposed to synthesize novel views of human geometry and appearance. While these methods achieved high-quality results, they generally require multi-view data collected with calibrated cameras and have high computational cost, notably during the inference/test time.

In this paper, we propose a carefully designed architecture for human neural rendering, leveraging the new possibilities offered by differentiable rendering techniques~\cite{loper2014opendr,liu2019softras,ravi2020pytorch3d,zhang2020image}. We present an end-to-end learning method that i) does not require data capture with calibrated systems, ii) is computationally efficient in test time, and iii) leverages DR and adversarial training to improve the estimation capabilities of fully controllable realistic 3D texture-mapped models of humans. 

%for estimating a full 3D textured model of people in this paper we techniques\cite{loper2014opendr,liu2019softras,ravi2020pytorch3d,mildenhall2020nerf} open several research potential by allowing reas the rendering process. In this p Inspired by the capability of GCNs and GANs, and the promising idea of applying these approaches together with differentiable rendering~\cite{zhang2020image}, our proposed methodology improves the capability of estimate controllable realistic 3D texture-mapped models of humans.

\section{Methodology}

\begin{figure*}[t!]
	\centering
	\includegraphics[width=0.9\linewidth]{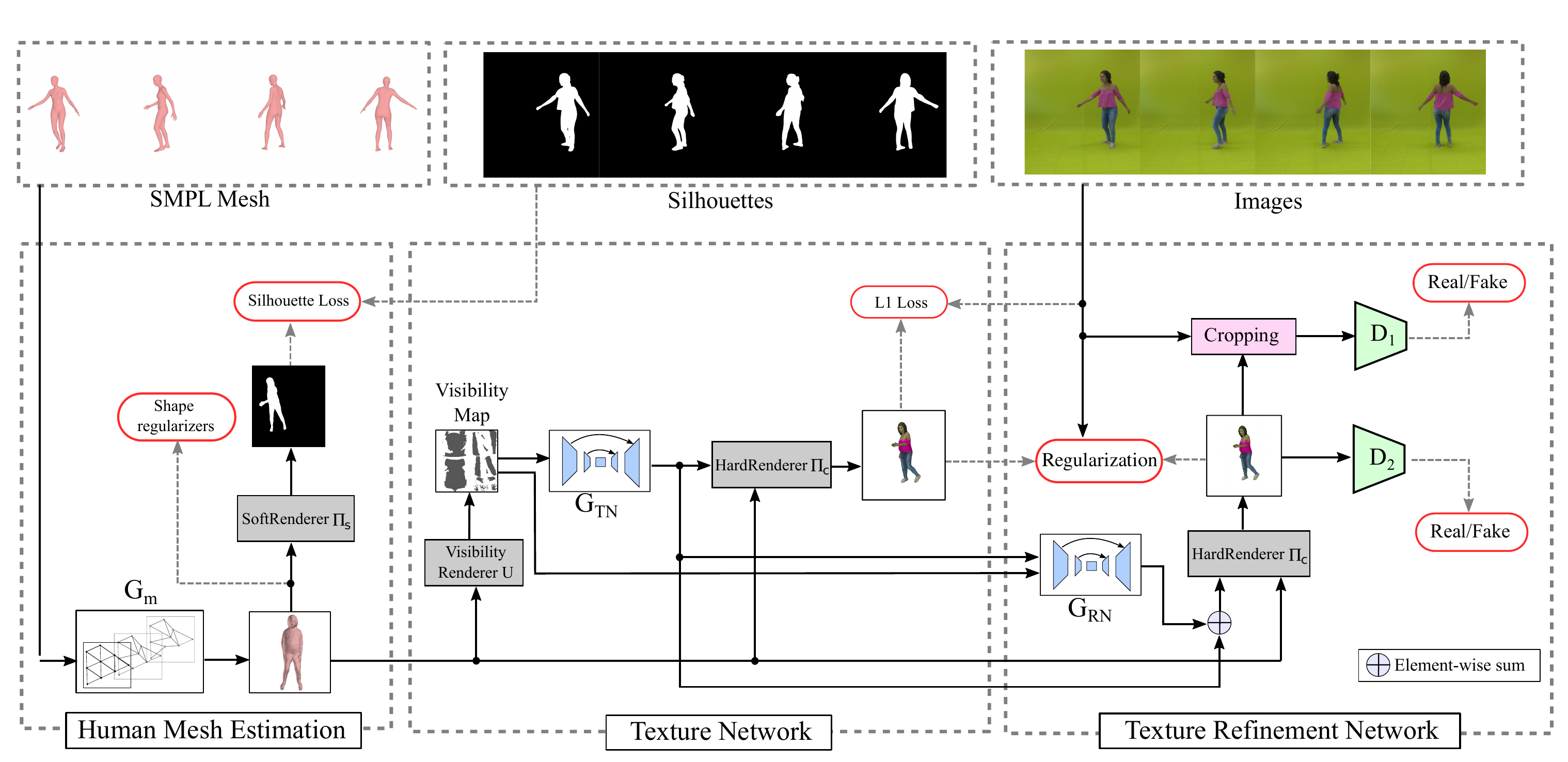}
	\caption{\textbf{Human synthesis architecture.} Our architecture has three main networks: \textbf{a)} the \textit{Human Mesh Estimation} that comprises a three-stage GCN and learns to fit and deform the mesh based on rendered silhouettes and shape regularizers; \textbf{b)} the \textit{Texture Network}, a CNN that is trained conditioned on the visibility map generated from the deformed mesh to create a coarse texture by rendering and optimizing on the $l_{1}$ norm; \textbf{c)} the \textit{Texture Refinement Network}, a second CNN that is conditioned on the visibility map and the coarse texture to generate the detailed texture map of the person.} % It is trained in an adversarial manner to generate high frequency detail textures by rendering. A regularization loss based on the $l_{1}$ norm is used to avoid artifacts in the rendered images.} 	
	\label{fig:method}\vspace*{-0.2cm}
\end{figure*}

%In the training phase, our model receives a sequence of images of an actor performing actions, 2D poses as well as its segmented silhouettes on the real images, as shown in the top of Figure~\ref{fig:method}. 

In order to generate deformable 3D texture-mapped human models, our end-to-end architecture has three main components to be trained during the rendering. The first component models local deformations on the human 3D body shape extracted from the images using a three-stage GCN. 
%The GCN is trained to fit and deform a 3D mesh of the human body into the silhouette of its rendered image (the rendered 2D projection). 
In the second component, a CNN is trained to estimate the human appearance map. Similar to the GCN, the CNN is trained in a self-supervised regime using the gradient signals from the differentiable renderer. Finally, the third component comprises an adversarial regularization in the human appearance texture domain to ensure the reconstruction of photo-realistic images of people.  

All frames of a person are used to train the mesh and the texture networks. The frames provide different viewpoints of the person to build his/her texture map and to estimate the mesh deformation according to the motion. We apply a stage-wise training strategy. First, we train the mesh model using the silhouettes, which produce a texture-less human model. In the sequence, we freeze the mesh network and train a global texture model. Then, we freeze both the mesh model and the global texture model and learn the texture refinement generator and the discriminators parameters. In the inference time, we feed our architecture with generic meshes parameterized by the SMPL model, and then to create a refined mesh and a detailed texture map to properly represent the person's shape and appearance. Figure~\ref{fig:method} outlines these components and their relations during the training phase and Figure~\ref{fig_method2} shows these components during the inference phase.

%\subsection{Shape and Appearance Representation}

%Although it is relatively common existence of applications that obtain static 3D character models through images, it requires additional effort to create an animated virtual character. In addition to the 3D model needing to be equipped with a skeleton hierarchy and appropriate skin weights, it is also necessary to fit a garment model into a human model in a variety of poses.

%In order to tackle these difficulties, we propose a shape and appearance representation of the human body that explores the global and local information of the human body. We encode the global information of the human body using the SMPL model parametrization~\cite{Loper_2015}, which is composed of a learned human shape distribution $\mathcal{M}$, $24$ 3D joint angles ($\boldsymbol{\theta} \in \mathbb{R}^{72}$ defining 3D rotations of the skeleton joint tree), and shape coefficients $\boldsymbol{\beta} \in \mathbb{R}^{10}$ that model the proportions and dimensions of the human body. While the global information provided by SMPL parametrization enables global control of human poses that can be used to retargeting human poses, they do not encode the fine details of the human body geometry (local information). In order to characterize the local information, we add a set of offsets to the mesh produced by the SMPL parametrization. Finally, we represent the appearance of the person as a texture map in a fixed UV space that can be applied to the refined mesh, \ie the SMPL mesh with offsets.

\subsection{Human Mesh Estimation}

\paragraph{Shape and pose representation.} We adopt a shape representation that explores the global and local information of the human body. We encode the global information using the SMPL model parametrization~\cite{Loper_2015}, which is composed of a learned human shape distribution $\mathcal{M}$, $24$ 3D joint angles $\boldsymbol{\theta} \in \mathbb{R}^{72}$ (defining 3D rotations of the skeleton joint tree), and shape coefficients $\boldsymbol{\beta} \in \mathbb{R}^{10}$ that model the proportions and dimensions of the human body. We used SPIN~\cite{kolotouros2019spin} to regress the SMPL parameters due to its trade-off between accuracy and efficiency. While the global information provided by SMPL parametrization enables global control of human poses, they do not encode the fine details of the human body geometry (local information). We then characterize the local information by adding a set of offsets to the mesh produced by the SMPL parametrization from the GCN mesh network. 

\vspace*{-0.35cm}
\paragraph{Mesh Refinement Network.}

%In this paper, we use the SPIN~\cite{kolotouros2019spin} as estimator to camera and SMPL parameters (global information) due to its good trade-off between accuracy and efficiency. 

After computing the global human shape and pose information $P = \text{SMPL}(\boldsymbol{\theta},\boldsymbol{\beta})$, we model local deformations on the mesh with the GCN Mesh Network $G_m$. Since the SMPL parametrization only provides a coarse 3D shape, and it cannot accurately model fine structures like clothes, we feed the mesh network with the initial SMPL mesh to refine its vertex positions with a sequence of refinement stages. The network is designed to learn the set of offsets to the SMPL mesh to increase the realism of the generated views. 
Drawing inspiration from the network architecture of~\cite{Gkioxari_2019_ICCV}, our refinement network is composed of three blocks with six graph convolutional layers with intermediate features of size $128$. Each block in our mesh network performs four operations: \vspace*{-0.2cm}

%vertex normal computation that extracts mesh features for vertices; graph convolution, which propagates information along mesh edges; vertex refinement, which computes  vertex offsets; and vertex re-projection, which updates vertex positions. Each layer of the network maintains a 3D position $v_i$ and a feature vector $f_i$ for each mesh vertex.
\begin{itemize}
	\item {\it Vertex normal computation.} This operation computes the normal surface vector for each vertex as being the weighted sum of the normals of faces containing the vertex, where the face areas are used as the weights. The resulting normal is assigned as the node feature ${f_i}$ for the vertex $v_i$ in the GCN.\vspace*{-0.2cm}
	\item {\it Graph convolution.} This convolutional layer propagates information along mesh edges using a aggregation strategy. Similar to~\cite{Gkioxari_2019_ICCV}, given the input vertex feature ${f_i}$, the layer updates the feature as ${f'_i = \text{ReLU}(W_0f_i + \sum_{j \in \mathcal{N}}W_1f_j)}$, where $W_0$ and $W_1$ are learned weighting matrices, and $\mathcal{N}(i)$ gives the i-th vertex’s neighbors in the mesh. \vspace*{-0.2cm}
	\item {\it Vertex refinement.} To improve the quality of the vertex position estimation, this operation computes vertex offsets as
	$u_i  = \tanh(W[f'_i ; f_i ])$. $W$ is a learned weighting matrix. \vspace*{-0.2cm}
	\item {\it Vertex refinement clamping.} To avoid strong deformations (large $||u_i||_2$), we constraint and bound the position update of each vertex $v_i$ as ${v'_i  = v_i + \min(\max(u_i,-K(v_i)),K(v_i))}$, where $K(v)$ is the 3D update bounds allowed to the vertex $v$, depending on the body part it belongs to. Each body part, \eg, face, footprints, hands, head, torso, arms, feet, \etc, have predefined bound thresholds. 
	%The correspondence between each vertex and body part threshold. 
	This operation ensures that the offsets do not exceed the threshold defined to that body part, and that the refinement of the mesh geometry do not affect the body's topology.%\vspace*{-0.2cm}
	
\end{itemize}

%Each of the three stages of the mesh network produces an intermediate mesh that is further refined by the next stage. 

%\begin{figure}[t!]
% 	\includegraphics[width=1.0\linewidth]{fig/fig2.png}
% 	\caption{\textbf{Mesh Refinement Network}.}
% 	
% 	\label{fig:mesh_network}
% \end{figure}
\vspace*{-0.7cm}
\paragraph{Loss function.}

For learning the mesh refinement, our model exploits two differentiable renderers that emulate the image formation process. Techniques such as~\cite{liu2019softras,ravi2020pytorch3d} enable us to invert such renderers and take the advantage of the learning paradigm to infer shape and texture information from the 2D images. 

%The loss train to minimize the differences between the image silhouette $\hat{I}_s$ of the human body obtained by rendering the 3D human model into the image by \textit{SoftRenderer}, a neural renderer,  and $I_s$ the human silhouette extracted from the input video. By inverting the renderer, we can define the loss of the Mesh Network as:

During the training, the designed loss minimizes the differences between the image silhouette extracted from the input real image $I_s$ and the image silhouette $\hat{I}_s$ of the human body obtained by rendering the deformed 3D human model $M$ into the image by \textit{SoftRenderer}, a differentiable renderer $\Pi_s(M)$. \textit{SoftRenderer} is a differentiable that synthesises the silhouette of the actor. We can define the loss of the Mesh Network $G_m$ as:
\begin{equation}
\mathcal{L}_m = \lambda_{gl}\mathcal{L}_{gl} + \lambda_{gn}\mathcal{L}_{gn} + \mathcal{L}_s,
\end{equation}
\noindent where $\mathcal{L}_{gl}$ and $\mathcal{L}_{gn}$ regularize the Laplacian and the normals consistency of the mesh respectively~\cite{ravi2020pytorch3d}, $\lambda_{gl}$ and $\lambda_{gn}$ are the weights for the geometrical regularizers, and 
\begin{equation}
\mathcal{L}_s = 1 - \frac{\left \| \hat{I}_s\otimes I_s \right \|_1}{\left \| (\hat{I}_s + I_s) - \hat{I}_s\otimes I_s \right \|_1},
\end{equation} 
\noindent is the silhouette loss proposed by Liu~\etal~\cite{liu2019softras}, where ${\hat{I}_s = \Pi_s(M)}$, $M = G_m(P)$ is the refined body mesh model, and $\otimes$ is the element-wise product. %, and $\left \| \cdot \right\|_1$ is the $l_1$ norm.

% silhouette loss $\mathcal{L}_s$ as being $\mathcal{L}_s = 1 - \frac{\left \| \hat{I}_s\otimes I_s \right \|_1}{\left \| (\hat{I}_s + I_s) - \hat{I}_s\otimes I_s \right \|_1} $, where $\otimes$ is the element-wise product. 

% The loss function used to train our Mesh Network is a linear combination of three losses: the silhouette loss $\mathcal{L}_s$~\cite{liu2019softras}, and two geometry losses $\mathcal{L}_{gl}$ and $\mathcal{L}_{gn}$ that regularizes the Laplacian and the Normals consistency of the mesh respectively~\cite{ravi2020pytorch3d}. 

% The final loss is a weighted sum of the three losses:
% \begin{equation}
% \mathcal{L}_m = \mathcal{L}_s + \lambda_1\mathcal{L}_{gl} + \lambda_2\mathcal{L}_{gn}
% \end{equation}
% \noindent where $\lambda_1$, and $\lambda_2$ are the weights for the geometry losses. 

\subsection{Human Texture Generation}

We represent the appearance of the human model as a texture map in a fixed UV space that is applied to the refined mesh, in our case, the SMPL mesh with offsets. Our full pipeline for texture generation is depicted in Figure \ref{fig:method}-b-c and consists of two stages: a coarse texture generation and then texture refinement. In an initial stage, given the refined 3D meshes of the actor $M$, the Texture Network $G_{TN}$ learns to render the appearance of the actor in the image. This texture is also further used to condition and regularize the Texture Refinement Network $G_{RN}$ in the second stage. 

%Second, to obtain texture details we train a texture refinement network in an adversarial manner. The two parts are trained separately and allow us to force the global consistency of the texture.

\vspace*{-0.35cm}
\paragraph{Texture Network.}

We start estimating a coarse texture map with a U-Net architecture~\cite{pix2pix2017}. %, which has shown great success in image denoising, deblurring, and style transfer applications. 
The input of the network is a visibility map $x_v$ and it outputs the texture map $x_p$. The visibility map indicates which parts in the texture map must be generated to produce a realistic appearance for the refined mesh. The visibility map is extracted by a parametric function $x_v = U(M)$ that maps points of refined mesh $M$ with positive dot product between the normal vector and the camera direction vector to a point $x_v$ in the texture space. We implement $U$ as a render of the 2D UV-map considering only faces with positive dot product between the normal vector and the camera direction vector. Thus, the network can learn to synthesize texture maps on demand focusing on the important parts for each viewpoint. 

Figure~\ref{fig:method}-b shows a schematic representation of the Texture Network training. The \textit{HardRenderer} $\Pi_c(M,x_p)$, represents the colored differentiable renderer that computes the output coarse textured image $\hat{I}$, of model $M$ and texture map $x_p$. In our case, $\hat{I} = \Pi_c(M,G_{TN}(U(M))$. Conversely to the \textit{SoftRenderer}, the differentiable \textit{HardRenderer} is used to propagate the detailed human texture appearance information (color). Specifically, we train the Texture Network to learn to generate a coarse texture by imposing the $l_1$ norm in the person's body region of the color rendered image as: 
%\begin{equation}\label{eq:global_text_loss}
%\mathcal{L}_{pt} = \frac{\left \| (\hat{I} -  I )\otimes ( \hat{I}_s \otimes I_s) ) \right \|_1}{\left \| ( \hat{I}_s \otimes I_s) \right \|_1}.
%\end{equation}
%\noindent where $\hat{I}$ is the output image of the rendering, $I$ the real image in the training set, $\hat{I}_s$ is the rendered silhouette and $I_s$ is the extracted silhouette from the the real image $I$.

\begin{equation}\label{eq:global_text_loss}
\mathcal{L}_{pt} = {\left \| (\hat{I} -  I )\otimes B  \right \|_1}/{\left \|  B \right \|_1},
\end{equation}
\noindent where $I$ is the real image in the training set and $B$ is the union of the visibility masks and real body regions.

%\new{In there a relationship between equations 3 and 4? They seem to be several terms in common, maybe simplify notation $\hat{I}_s \otimes I_s$ is appearing in all equations...}

%The Figure~\ref{fig:method}-b shows a schematic representation of the  training regime of the \textbf{Texture Network}, where the \textit{HardRenderer}, $\Pi_c(M,x_p)$, represents the colored differentiable renderer that computes the output coarse textured image $\hat{I}$, \ie, $\hat{I} = \Pi_c(M,G_{TN}(U(M))$. Conversely to the \textit{SoftRenderer}, the differentiable \textit{HardRenderer} is used to propagate the detailed human texture appearance information (color).

%\paragraph{Texture refinement.} We further improve the coarse texture to represent finer details. For that, we design the Texture Refinement Network $G_{RN}$ to conditioning the generation of the final texture map on a coherent output of the Texture Network $G_{TN}$ and the visibility map. %as illustrated in detail in~Figure~\ref{fig:texture_network}. 

%\paragraph{Texture refinement.} We further improve the coarse texture to represent finer details. For that, we design the Texture Refinement Network $G_{RN}$ which learns the details of the texture produced by the Texture Network $G_{TN}$. The Network is conditioned in the coarse texture input and generates a finer texture, which is added to the initial texture.
\vspace*{-0.35cm}
\paragraph{Texture refinement.} We further improve the coarse texture to represent finer details. For that, we design the Texture Refinement Network $G_{RN}$ to condition the generation of a new texture map from the coarse texture, on a coherent output of the Texture Network $G_{TN}$ and the visibility map. %as illustrated in detail in~Figure~\ref{fig:texture_network}. 

%\begin{figure}[t!]
% 	\includegraphics[width=1.0\linewidth]{fig/texture_network.pdf}
% 	\caption{\textbf{Texture Refinement Network}.}
% 	\label{fig:texture_network}
% \end{figure}

In our adversarial training, the Texture Refinement Network acts as the generator network $G_{RN}$ and engages in a minimax game against two task-specific discriminators: the Face Discriminator $D_1$ and Image Discriminator $D_2$. The generator is trained to synthesize texture maps in order to fool the discriminators which must discern between ``real'' images and ``fake'' images, where ``fake'' images are produced by the neural renderer using the 3D texture-mapped model estimated by the Mesh and Texture Networks. While the discriminator $D_1$ sees only the face region, the Image Discriminator $D_2$ sees the whole image. 

These three networks are trained simultaneously and drive each other to improve their inference capabilities, as illustrated in Figure~\ref{fig:method}-c. The Texture Refinement Network learns to synthesize a more detailed texture map to deceive the discriminators which in turn learn differences between generated outputs and ground truth data. The total loss function for the generator and discriminators for the rendering is then composed of three terms:
\begin{equation}
\begin{aligned}
\min\limits_{G_{RN}} ( \max\limits_{D_1}\mathcal{L}_{GAN}(G_{RN},D_1) + \\ \max\limits_{D_2}\mathcal{L}_{GAN}(G_{RN},D_2) + \mathcal{L}_{r}(G_{RN}) ),
\end{aligned}
\end{equation}
\noindent where both $\mathcal{L}_{GAN}$ terms address the discriminators and $\mathcal{L}_{r}$ is a regularization loss to reduce the effects from outlier poses. Each adversarial loss is designed as follows:
\begin{equation}
\begin{aligned}
\mathcal{L}_{GAN}(G,D) = \mathbb{E}_{y}[\log D(y)] + \\ \mathbb{E}_{x_v,x_p}[\log (1 - D(\Pi_c(M,G(x_v,x_p))))],
\end{aligned}
\end{equation}
\noindent where $x_v$ is the visibility map, $x_p$ is the output of the Texture Network, $M$ is the refined mesh, and $y$ is the corresponding segmented real image $I \otimes B$.

%We also include in the loss with a more traditional loss $L_1$. 
Finally, to reduce the effects of wrong poses, which causes mismatches between the rendered actor silhouette and silhouette of the real actor, we also add a regularization loss to prevent the GAN to apply the color of the background into the human texture. The first term of the regularization loss acts as a reconstruction of the pixels by imposing the $l_1$ norm in the person's body region and the second term enforces eventual misaligned regions to stay close to the coarse texture:
%\begin{equation}
%\begin{aligned}
%\mathcal{L}_r = \alpha_1\frac{\left \| (y - \Pi(M,G(x_v,x_p)))\otimes ( \hat{I}_s \otimes I_s)\right \|_1}{\left \| ( \hat{I}_s \otimes I_s)\right \|_1} + \\
%\alpha_2\frac{\left \| (\Pi(M,x_p) - \Pi(M,G(x_v,x_p)))\otimes (\hat{I}_{sb} - (\hat{I}_{sb} \otimes I_s)) \right \|_1 }{\left \| \hat{I}_{sb} - (\hat{I}_{sb} \otimes I_s) \right \|_1},
%\end{aligned}
%\end{equation}
%\noindent where $\alpha_1$ and $\alpha_2$ are the weights, $\otimes$ is the element-wise product, $\hat{I}_s$ is the rendered silhouette, $I_s$ is the estimated silhouette and $I_{sb}$ is the the rendered silhouette without the face region. 
\begin{equation}
\begin{aligned}
\mathcal{L}_r = \alpha_1{\left \| (I - \hat{I}^{RN})\otimes B \right\|_1}/{\left \| B \right \|_1} + \\
\alpha_2{\left \| (\hat{I}^{TN} - \hat{I}^{RN})\otimes C\right \|_1 }/{\left \| C \right \|_1},
\end{aligned}
\end{equation}
\noindent where $\alpha_1$ and $\alpha_2$ are the weights, $\hat{I}^{TN}$ is the rendered image using the coarse texture, $\hat{I}^{RN}$ is the rendered image using the refined texture, and $C$ is the misaligned regions without the face region, \ie, the image region where the predicted silhouette and estimated silhouette are different.

\section{Experiments and Results}

%human retargeting and appearance transfer and has a diversity of actors with different body shapes, gender, clothing styles and sizes. The set of movements performed by each actor were chosen to be representative for the task of human retargeting and appearance transfer, presenting different levels of difficulty that aims to test methods generalization in a set of data that has not been presented in the training regimen. The movements are of "pick up a box", "spinning", "jump", "walk", "shake hands", "touch a cone", "pull down" and "fusion dance".
% 

\paragraph{Datasets and baselines.} 
%We start providing a short description of the datasets used for training and evaluation. 
For the training of both texture models and the mesh human deformation network we considered four-minute videos provided by~\cite{gomes2020arxiv}, where the actors perform random movements, allowing the model to get different views from the person in the scene. We use the SMPL model parameters calculated by SPIN~\cite{kolotouros2019spin} and the silhouette image segmented by MODNet~\cite{MODNet} for each frame of the video. In the evaluation/test time of our 3D human rendering approach, and to provide comparisons to related methods, we conducted experiments with publicly available videos used by Chan~\etal~\cite{chan2018dance}, Liu~\etal~\cite{lwb2019}, and Gomes~\etal~\cite{gomes2020arxiv} as the evaluation sequences. %The later enables the evaluation of methods to perform neural rendering of people in different contexts. %\paragraph{Comparison with Other Methods.}

We compare our method against five recent methods including V-Unet~\cite{Esser_2018_CVPR}, Vid2Vid~\cite{wang2018vid2vid}, EBDN~\cite{chan2018dance}, Retarget~\cite{gomes2020arxiv} and the Impersonator~\cite{lwb2019}. V-Unet is a famous method of image-to-image translation that uses conditional variational autoencoders to generate images based on a 2D skeleton and an image from the target actor. The approach Retarget is an image-based rendering method based on image rendering designed to perform human retargeting. Impersonator, Vid2Vid, and EBDN are generative adversarial models trained to perform human neural rendering. 
\vspace*{-0.35cm}
\paragraph{Metrics and evaluation protocol.} We adopted complementary metrics to evaluate the quality of the approaches to asset different aspects of the generated images such as structure coherence, luminance, contrast, perceptual similarity~\cite{zhang2018perceptual}, temporal, and spatial coherence. The metrics used to perform quantitative analysis are SSIM~\cite{Wang04imagequality}, LPIPS~\cite{zhang2018perceptual}, Mean Square Error (MSE), and Fréchet Video Distance (FVD)~\cite{unterthiner2019accurate}. Following the protocol proposed by~\cite{gomes2020arxiv}, we executed all the methods in the motion sequences and transfered them to the same background. This protocol allows us to generate comparisons with the ground truth and compute the metrics for all the generated images with their respective real peers. Then, we group the values generated by the metrics in two ways: Motion Types and Actors. In the first one, for each metric, we calculate the average values of all actors making a motion sequence (\eg, ``spinning"), while in the second one, we calculate the average values of all movements performed by a given actor. This grouping allows us to analyze the capability of the methods to render the same actor performing various movements with different levels of difficulty and also to compare their capacity to render actors with different morphology performing the same movement.
\vspace*{-0.35cm}
\paragraph{Implementation details.} %We use PyTorch3D~\cite{ravi2020pytorch3d} implementation of differentiable rendering and GCN operators. 
We trained our body mesh refinement network for $20$ epochs with batch size of $4$. We used AdamW~\cite{AdamW} with parameters $\beta_{1} = 0.5$, $\beta_{2} = 0.999$, weight decay = $1\times10^{-2}$, and learning rate of $1\times10^{-4}$ with a linear decay routine to zero starting from the middle of the training. We empirically set $\lambda_{1}$ and $\lambda_{2}$ to $1.0$ and $0.5$, respectively. In the Vertex refinement clamping component, we defined the set of thresholds as follows: $K \in$ \{face = $0.0$; footprints = $0.0$; hands = $0.0$; head = $0.04$; torso = $0.06$; arms = $0.02$; forearms = $0.04$; thighs = $0.04$; calves = $0.03$; feet = $0.02$\} meters. All the training and inference were performed in a single Titan XP ($12$ GB), where the GCN mesh model and the human texture networks took around $6$ and $20$ hours per actor, respectively. The inference time takes $92$ ms per frame ($90$ ms in the GCN model deformation and $2$ ms in the texture networks).

Due to remarkable performance of pix2pix~\cite{pix2pix2017} in  synthesizing photo-realistic images, we build our Texture Network upon its architecture. The optimizers for the texture models were configured as the same as the Mesh Network, except for the learning rates. The learning rate for the whole body and face discriminators, the global texture and refinement texture generators were set as $2\times10^{-5}$, $2\times10^{-5}$, $2\times10^{-3}$, and $2\times10^{-4}$, respectively. The parameters of the texture reconstruction was set to $\alpha_{1} = 100$ and the regularization as $\alpha_{2} = 100$. We observed that smaller values led to inconsistency in the final texture. For the training regime, we used $40$ epochs with batch size $8$. The global texture model was trained separately from the other models for $2{,}000$ steps, then we freeze the model, the texture refinement generator and the discriminators were trained.

\begin{figure}[t!]
	\includegraphics[width=1\linewidth]{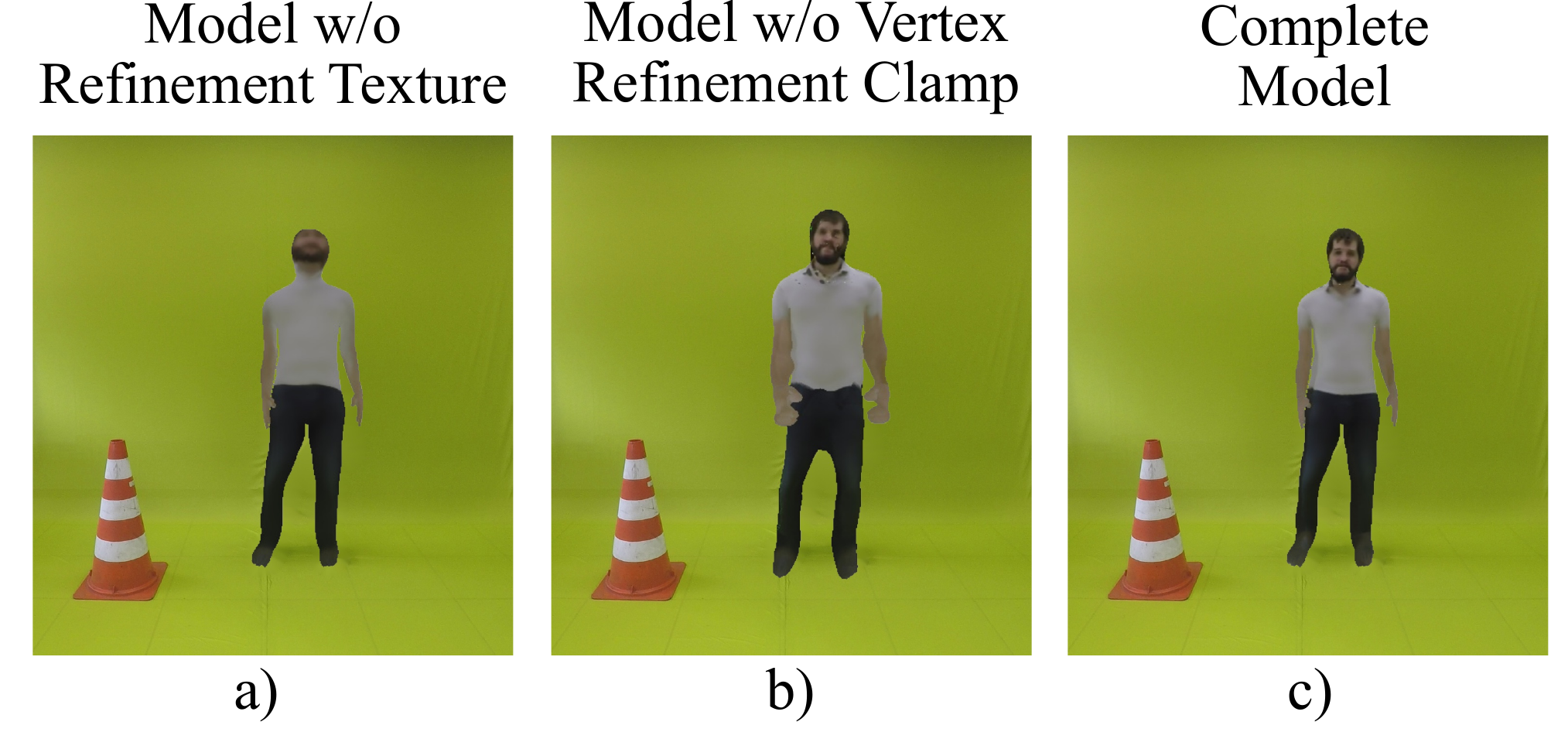}\vspace*{-0.4cm}
	\caption{\textbf{Ablation study.} a) Results of the texture training without the refinement stage; b) Model without the Vertex Refinement Clamp layer. We observe an excessive growth of the mesh without update thresholds. The texture produced lacks details and even could not preserve the actor's face; c) shows the results for our complete model.}
	\label{fig:ablation_figure}
\end{figure}

\begin{table}[t!]
	\centering
	\caption{{\bf Ablation study}. SSIM, LPIPS, MSE, and FVD comparison by motion types. Best in bold.}
	\label{table:ablation_result}
	\resizebox{\columnwidth}{!}{%
		
		\begin{tabular}{clrrrr}
			\toprule
			\multirow{3}{*}{\bf Method}
			&  \multicolumn{1}{c}{}
			&  \multicolumn{4}{c}{{\bf Metrics}}\\ \cmidrule{3-6}
			&  \multicolumn{1}{c}{} 
			&  {\centering SSIM$^1$}
			&  {\centering LPIPS$^2$}
			&  {\centering MSE$^2$}
			&  {\centering FVD$^2$} \\ \midrule
			
			& \hspace*{-1.9cm} Texture Refinement Removal & $\textbf{0.869}$ & $0.136$ & $262.39$ & $795.15$ \\
			
			& \hspace*{-1.9cm} Vertex Refinement Clamping & $0.866$ & $0.142$ & $288.04$ & $829.60$\\
			
			& \hspace*{-1.9cm} Complete Model & $0.868$ & $\textbf{0.134}$ & $\textbf{259.79}$ & $\textbf{769.54}$  \\
			
			& & \multicolumn{2}{c}{\scriptsize{$^1$\textit{Higher is better}}} & \multicolumn{2}{c}{\scriptsize{$^2$\textit{Lower is better}}} \\	
			\bottomrule
			\vspace{-0.8cm}
		\end{tabular}
		
	}
\end{table}

\vspace*{-0.35cm}
\paragraph{Ablation Study.}

We evaluate the contributions of different parts of the method to the overall view synthesis performance. We investigated the benefits from the Vertex refinement clamping component in the Mesh Refinement Network (MRN) and the use of adversarial training in the texture generation. For the first experiment, we removed the vertex refinement thresholds, letting the mesh grow loosely. All other steps of texture training were maintained. Table~\ref{table:ablation_result} shows that the performance dropped drastically when compared to our original model. A qualitative analysis of the results in Figure~\ref{fig:ablation_figure}-b demonstrates that removing the Vertex refinement clamping component led to strong wrong deformations in the hands and feet, \ie, the regions with higher pose estimation errors.% In addition, setting the threshold to zero in some vertex helps to constraint the global mesh position, preventing the displacement of the entire mesh together. 

%the mesh model optimized even further the loss function, 

In the adversarial training analysis, we maintained the original Mesh Refinement Network and removed the Texture Refinement Network and its discriminators, training only the Global Texture Network using Equation~\ref{eq:global_text_loss}. Figure~\ref{fig:ablation_figure}-a shows the texture quality of the models trained with and without the adversarial regime. After removing the GAN the model could not generate textures with fine details, producing blurry results. This result is also reported in the metrics of Table~\ref{table:ablation_result}, where we show the average values calculated from all motion sequences in the test data in which the model without GAN is outperformed in all results besides SSIM. This result is coherent, since SSIM is based on low-level image features, and blurred textures can lead to higher SSIM scores.

\begin{table*}[t!]
	\centering
	\caption{{\bf Comparison with state-of-the-art human neural rendering}. SSIM, LPIPS, MSE, and FVD comparison by motion and actors types. Best in bold.}
	\label{table:metrics_result}
	\resizebox{0.92\linewidth}{!}{%
		
		\begin{tabular}{@{}clrrrrrrrrrrrrr@{}}
			\toprule %\hline
			\multirow{3}{*}{\bf Metric} & \multirow{3}{*}{\bf Method} & \multicolumn{8}{c}{{\bf Motion type}} & \phantom{a} & \multicolumn{4}{c}{{\bf Actor type}}\\ \cmidrule{3-10} \cmidrule{12-15} 
			& \multicolumn{1}{r}{} 
			&  {\centering jump}
			&  {\centering walk}
			&  {\centering spinning}
			&  {\centering shake hands}
			&  {\centering cone}
			&  {\centering fusion dance}
			&   {\centering pull down}
			&  {\centering box}
			& \phantom{a}
			&  {\centering S1} 
			&  {\centering S2} 
			&  {\centering S3}  
			&  {\centering S4} \\ \midrule
			
			\multirow{6}{*}{\rotatebox[origin=c]{90}{\parbox[c]{1.5cm}{\centering SSIM$^1$}}} 
			
			& EBDN~\cite{chan2018dance} & $0.878$ & $0.880$ & $0.855$ & $0.859$ & $0.878$ & $0.820$ & $0.857$  & $0.858$ &  \phantom{a}
			& $0.867$ & $0.898$  & $0.844$ & $0.834$  \\
			
			& Imper~\cite{lwb2019} & $0.877$ & $0.880$ & $0.852$ & $0.859$ & $0.877$ & $0.816$ & $0.855$ & $0.856$ & \phantom{a} 
			& $0.867$ & $0.896$  & $0.842$ & $0.831$\\
			
			& Retarget~\cite{gomes2020arxiv} & $0.881$ & $0.885$ & $0.855$ & $0.860$ & $0.879$ & $0.820$ & $0.861$ & $0.869$ & \phantom{a}
			& $0.872$ & $0.902$  & $0.846$ & $0.834$  \\
			
			& Vid2Vid~\cite{wang2018vid2vid} & $0.880$ & $0.884$ & $0.856$ & $0.858$ & $0.878$ & $0.821$ & $0.859$ &  $0.866$ &  \phantom{a}
			& $0.868$ & $0.901$  & $0.848$ & $0.835$ \\
			
			& V-Unet~\cite{Esser_2018_CVPR} & $0.870$ & $0.871$ & $0.843$ & $0.847$ & $0.862$ & $0.797$ & $0.847$ &  $0.857$ &  \phantom{a}
			& $0.855$ & $0.886$  & $0.830$ & $0.826$   \\
			
			& Ours & $\textbf{0.884}$ & $\textbf{0.890}$ & $\textbf{0.860}$ & $\textbf{0.865}$ & $\textbf{0.885}$ & $\textbf{0.824}$ & $\textbf{0.866}$ & $\textbf{0.873}$ &  \phantom{a}
			& $\textbf{0.876}$ & $\textbf{0.908}$  & $\textbf{0.852}$ & $\textbf{0.838}$  \\
			
			\midrule
			
			\multirow{6}{*}{\rotatebox[origin=c]{90}{\parbox[c]{1.5cm}{\centering LPIPS$^2$}}} 
			
			& EBDN~\cite{chan2018dance} & $0.141$ & $0.122$ & $0.139$ & $0.138$ & $0.143$ & $0.215$ & $0.151$  & $0.170$ & \phantom{a}
			& $0.159$ & $0.145$  & $0.147$ & $0.159$   \\
			
			& Imper~\cite{lwb2019} & $0.151$ & $0.134$ & $0.151$ & $0.151$ & $0.155$ & $0.239$ & $0.168$ & $0.184$ & \phantom{a} 
			& $0.161$ & $0.165$  & $0.170$ & $0.171$  \\
			
			& Retarget~\cite{gomes2020arxiv} & $\textbf{0.125}$ & $0.099$ & $\textbf{0.130}$ & $0.131$ & $0.128$ & $\textbf{0.206}$ & $\textbf{0.131}$ & $\textbf{0.127}$ & \phantom{a}
			& $\textbf{0.133}$ & $0.129$  & $\textbf{0.133}$ & $\textbf{0.143}$  \\
			
			& Vid2Vid~\cite{wang2018vid2vid} & $0.131$ & $0.105$ & $0.126$ & $0.136$ & $0.133$ & $0.203$ & $0.142$ &  $0.133$ &  \phantom{a}
			& $0.148$ & $0.131$  & $0.129$ & $0.147$   \\
			
			& V-Unet~\cite{Esser_2018_CVPR} & $0.147$ & $0.132$ & $0.157$ & $0.161$ & $0.174$ & $0.243$ & $0.166$ &  $0.158$ & \phantom{a}
			& $0.184$ & $0.160$  & $0.166$ & $0.158$   \\
			
			& Ours & $0.127$ & $\textbf{0.097}$ & $\textbf{0.130}$ & $\textbf{0.130}$ & $\textbf{0.124}$ & $\textbf{0.206}$ & $0.132$ & $\textbf{0.127}$ &  \phantom{a}
			& $0.136$ & $\textbf{0.124}$  & $0.134$ & $\textbf{0.143}$   \\
			
			\midrule			
			
			\multirow{6}{*}{\rotatebox[origin=c]{90}{\parbox[c]{1.5cm}{\centering MSE$^2$}}} 
			
			& EBDN~\cite{chan2018dance} & $306.92$ & $269.58$ & $312.69$ & $312.12$ & $266.17$ & $463.79$ & $384.23$  & $361.57$ & \phantom{a}
			& $324.40$ & $314.10$  & $331.83$ & $368.19$  \\
			
			& Imper~\cite{lwb2019} & $313.43$ & $275.22$ & $344.94$ & $314.92$ & $267.39$ & $504.00$ & $404.79$ & $377.16$ & \phantom{a} 
			& $277.98$ & $328.07$  & $358.30$ & $436.57$\\
			
			& Retarget~\cite{gomes2020arxiv} & $237.16$ & $178.57$ & $286.86$ & $270.25$ & $237.86$ & $434.64$ & $301.66$ & $245.86$ & \phantom{a}
			& $\textbf{243.95}$ & $260.14$  & $294.88$ & $297.46$  \\
			
			& Vid2Vid~\cite{wang2018vid2vid} & $257.33$ & $206.42$ & $286.18$ & $332.09$ & $253.04$ & $452.77$ & $349.28$ &  $288.54$ &  \phantom{a}
			& $312.40$ & $274.80$  & $269.81$ & $356.26$   \\
			
			& V-Unet~\cite{Esser_2018_CVPR} & $295.04$ & $269.59$ & $354.33$ & $377.02$ & $328.68$ & $559.75$ & $417.00$ &  $346.13$ &  \phantom{a}
			& $381.77$ & $344.71$  & $362.63$ & $384.66$\\
			
			& Ours & $\textbf{231.20}$ & $\textbf{153.23}$ & $\textbf{278.75}$ & $\textbf{254.38}$ & $\textbf{218.76}$ & $\textbf{418.58}$ & $\textbf{286.02}$ & $\textbf{237.42}$ &  \phantom{a}
			& $247.10$ & $\textbf{236.49}$  & $\textbf{276.17}$ & $\textbf{279.42}$   \\
			
			\midrule
			\multirow{6}{*}{\rotatebox[origin=c]{90}{\parbox[c]{1.5cm}{\centering FVD$^2$}}} 
			
			& EBDN~\cite{chan2018dance} & $887.56$ & $273.00$ & $918.94$ & $\textbf{423.08}$ & $725.49$ & $952.22$ & $1,113.46$  & $853.26$ &  \phantom{a}
			& $791.98$ & $751.45$  & $\textbf{560.27}$ & $826.71$\\
			
			& Imper~\cite{lwb2019} & $1,770.31$ & $656.07$ & $1,531.64$ & $1,266.14$ & $1,051.42$ & $1,322.72$ & $1,440.94$ & $1,719.55$ &  \phantom{a} 
			& $1,270.41$ & $1,092.82$  & $1,395.81$ & $1,214.64$  \\
			
			& Retarget~\cite{gomes2020arxiv} & $1,119.50$ & $330.91$ & $\textbf{674.99}$ & $478.93$ & $767.68$ & $791.01$ & $\textbf{988.35}$ & $\textbf{760.33}$ &  \phantom{a}
			& $\textbf{715.00}$ & $\textbf{653.30}$  & $720.49$ & $\textbf{515.61}$  \\
			
			& Vid2Vid~\cite{wang2018vid2vid} & $\textbf{879.94}$ & $266.6$ & $1,085.49$ & $396.31$ & $790.79$ & $997.42$ & $997.96$ &  $1,069.85$ &  \phantom{a}
			& $778.53$ & $719.80$  & $762.46$ & $574.08$   \\
			
			& V-Unet~\cite{Esser_2018_CVPR} & $1,491.63$ & $845.44$ & $1,721.81$ & $1,257.20$ & $1,415.24$ & $1,712.93$ & $2,437.98$ &  $1,816.94$ & \phantom{a}
			& $2,239.14$ & $1,352.10$  & $1,856.78$ & $1,108.34$  \\
			
			& Ours & $1,114.43$ & $\textbf{233.81}$ & $1019.83$ & $542.24$ & $\textbf{614.88}$ & $\textbf{746.22}$ & $1010.24$ & $874.69$ & \phantom{a}
			& $881.49$ & $697.68$  & $718.21$ & $551.06$ \\

			& & \multicolumn{6}{c}{\scriptsize{$^1$\textit{Higher is better}}} & \multicolumn{6}{c}{\scriptsize{$^2$\textit{Lower is better}}} \\	
			\bottomrule
		\end{tabular}
		
	}
	
\end{table*}

\subsection{Results}

\paragraph{Quantitative comparison with state of the art.}
%human retargeting and appearance transfer and has a diversity of actors with different body shapes, gender, clothing styles and sizes. The set of movements performed by each actor were chosen to be representative for the task of human retargeting and appearance transfer, presenting different levels of difficulty that aims to test methods generalization in a set of data that has not been presented in the training regimen. The movements are of "pick up a box", "spinning", "jump", "walk", "shake hands", "touch a cone", "pull down" and "fusion dance".

\begin{figure}[t!]
	\vspace*{-0.3cm}
	\centering
	\includegraphics[width=.95\linewidth]{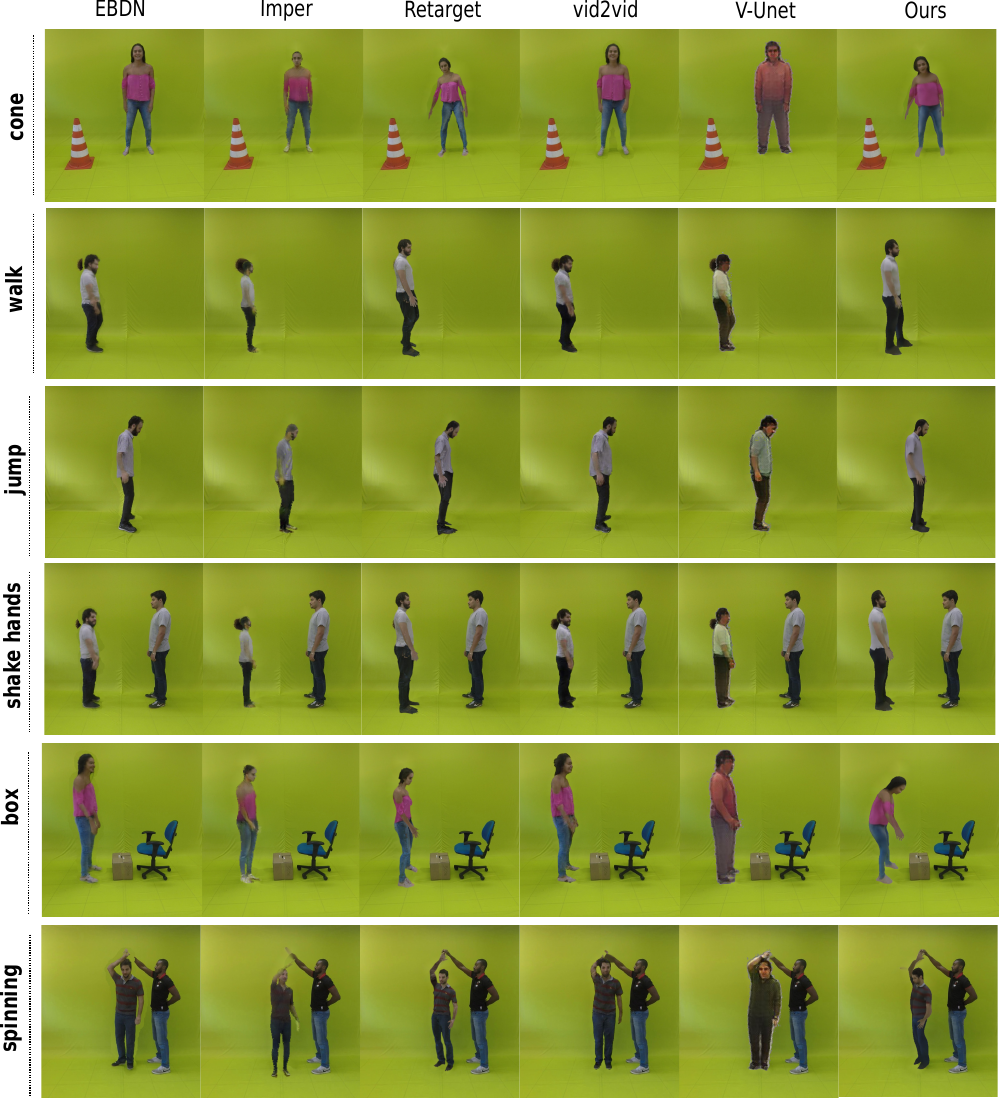}\vspace*{-0.2cm}
	\caption{\textbf{Qualitative comparison.} Movements with four different actors are represented in the rows. The results for each competitor are represented in the columns.}
	\label{fig:results_dataset}\vspace*{-0.3cm}
\end{figure}

\begin{figure}[t!]
	\vspace*{-0.3cm}
	\includegraphics[width=0.95\linewidth]{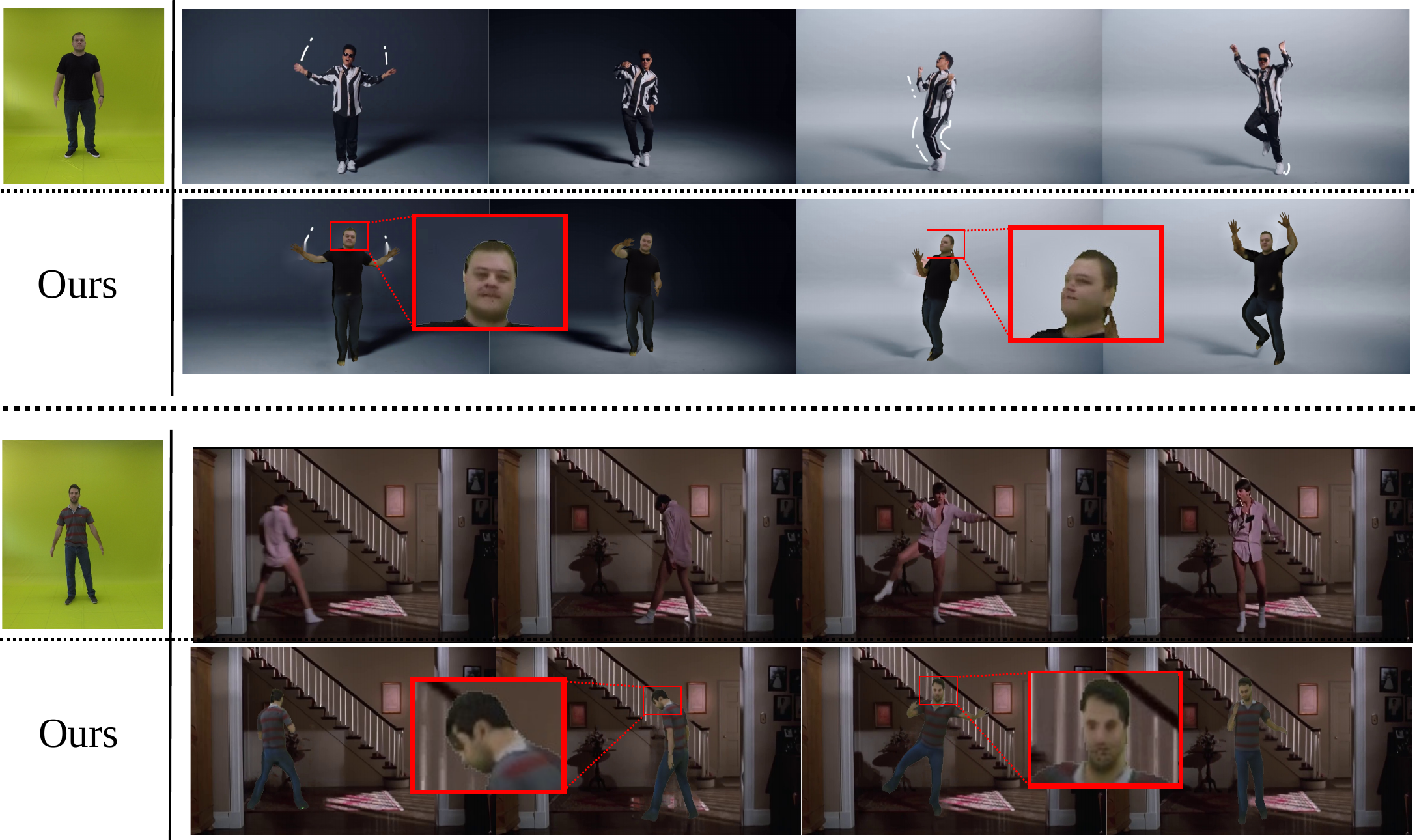}\vspace*{-0.2cm}
	\caption{\textbf{Human appearance transfer and animation with different scenes and camera setups}. The first line of each scene illustrates the original frames of the source video. On the second line is the transferred appearance of the animated virtual actor using our proposed method. The red squares highlight the face generation quality.}
	\label{fig:bruno-tom} \vspace*{-0.3cm}
\end{figure}

We performed the neural rendering for actors with different body shapes, gender, clothing styles, and sizes for all considered video sequences. The video sequences used in the actors' animations contained motions with different levels of difficulty, which aims to test the generalization capabilities of the methods in unseen data. %Some movements are of ``pick up a box", ``spinning", "jump", "walk" or "shake hands". %, including public dance videos also adopted in the works Chan \textit{et al.} \cite{chan2018dance} and Liu \textit{et al.} \cite{liu2019liquid}. 
Table~\ref{table:metrics_result} shows the performance for each method  considering all motion and actors types in the dataset.
%It is worth mentioning that  Everybody Dance Now and Vid2vid are dataset-specific methods, so we trained both methods on the dataset in order to make fair comparisons. 
We can see that our method achieves superior peformance as compared to the methods in most of the motion sequences and actor types considering the SSIM, LPIPS, MSE, and FVD metrics. These results indicate that our method is capable of deforming the mesh according to the shape of the given actors and then, rendering a texture optimized to fit the morphology of the person in the scene. Furthermore, our training methodology, that considers multiple views from the actor and the shape parameters, allows the generation of consistent rendering with less visual artifacts when the character is performing challenging movements, such as bending or rotating.
\vspace*{-0.35cm}
\paragraph{Qualitative visual analysis.} The visual inspection of synthesized actors also concur with the quantitative analysis. Figure \ref{fig:results_dataset} shows the best frames for each movement using four actors in the dataset. Our model and Retarget~\cite{gomes2020arxiv} are the only approaches capable of keeping the body scale of the authors along all scenes, while the other methods failed, in particular in the movements \textit{shake hands} and \textit{walk}. Besides generating coherent poses, our technique also generated more realistic textures in comparison to the other methods. Comparing the results of the movements \textit{jump} and \textit{spinning}, one can visualize some details as the shadow of the shirt sleeve of the actor and the shirt collar, respectively. Figure~\ref{fig:bruno-tom} illustrates a task of transferring the appearance in videos with different scenes and camera-to-actor translations and camera intrinsics. These results also demonstrate our capability of generating detailed face and body texture, producing a good fit to synthesize views of actors observed from different camera setups. 

\vspace*{-0.2cm}
\section{Conclusions}

In this paper we proposed a method that produces a fully 3D controllable representation of people from images. Our key idea is leveraging differentiable rendering, GCNs, and adversarial training to improve the capability of estimating realistic 3D texture-mapped models of humans. Taking advantages of both differentiable rendering and the 3D parametric model, our method is fully controllable, which allows controlling the human pose and rendering parameters. Furthermore, we have introduced a graph convolutional architecture for mesh generation that refines the human body structure information, which results in a more accurate human mesh model. The experiments show that the proposed method has superior quality compared to recent neural rendering techniques in different scenarios, besides having several advantages in terms of control and ability to generalize to furthest views. 

%Our results indicate the potential of differentiable rendering approaches have the potential to push forward the area.

%Yet we observed none of the existing methods obtain artifact-free results, which suggests the problem of synthesising realistic views of people in general contexts is still a challenging problem. On the other hand, our results indicate that neural networks and differentiable rendering approaches have the potential to push forward the area. \new{These final comments are super pessimist. We do not see then in the results section...}

\vspace{-0.3cm}

{ \paragraph*{Acknowledgments.} The authors would like to thank CAPES, CNPq, FAPEMIG, and Petrobras for funding different parts of this work. R. Martins was also supported by the French National Research Agency through grants ANR MOBIDEEP (ANR-17-CE33-0011), ANR CLARA (ANR-18-CE33-0004) and by the French Conseil Régional de Bourgogne-Franche-Comté. We also thank NVIDIA Corporation for the donation of a Titan XP GPU used in this research.}

\clearpage
\balance
{\small
\bibliographystyle{ieee_fullname}
\bibliography{references}
}

\end{document}